%% file: main_meta.tex
\documentclass{fairmeta}

\input{preamble_shared}

\usepackage{wrapfig}  

\renewcommand{\methodname}{\textbf{UniT}}

\title{UniT: Unified Multimodal Chain-of-Thought Test-time Scaling}

\author[1,2]{Leon Liangyu Chen}
\author[2]{Haoyu Ma}
\author[2]{Zhipeng Fan}
\author[2,3]{Ziqi Huang}
\author[2]{Animesh Sinha}
\author[2]{Xiaoliang Dai}
\author[2]{Jialiang~Wang}
\author[2]{Zecheng He}
\author[2]{Jianwei Yang}
\author[2]{Chunyuan Li}
\author[2]{Junzhe Sun}
\author[2]{Chu Wang}
\author[1]{Serena~Yeung-Levy}
\author[2]{Felix~Juefei-Xu}

\affiliation[1]{Stanford University}
\affiliation[2]{Meta Superintelligence Labs}
\affiliation[3]{Nanyang Technological University}

\date{\today}

\metadata[Paper]{\small{\url{https://ai.meta.com/research/publications/unit-unified-multimodal-chain-of-thought-test-time-scaling}}}
\correspondence{Leon Liangyu Chen at \email{liangyuc@stanford.edu}}

\abstract{
Unified models can handle both multimodal understanding and generation within a single architecture, yet they typically operate in a single pass without iteratively refining their outputs. Many multimodal tasks, especially those involving complex spatial compositions, multiple interacting objects, or evolving instructions, require decomposing instructions, verifying intermediate results, and making iterative corrections. While test-time scaling (TTS) has demonstrated that allocating additional inference compute for iterative reasoning substantially improves language model performance, extending this paradigm to unified multimodal models remains an open challenge.

We introduce \methodname, a framework for multimodal chain-of-thought test-time scaling that enables a single unified model to reason, verify, and refine across multiple rounds. \methodname~combines agentic data synthesis, unified model training, and flexible test-time inference to elicit cognitive behaviors including verification, subgoal decomposition, and content memory. Our key findings are: (1) unified models trained on short reasoning trajectories generalize to longer inference chains at test time; (2) sequential chain-of-thought reasoning provides a more scalable and compute-efficient TTS strategy than parallel sampling; (3) training on generation and editing trajectories improves out-of-distribution visual reasoning. These results establish multimodal test-time scaling as an effective paradigm for advancing both generation and understanding in unified models.
}

\begin{document}

\maketitle

\begin{figure}[!t]
\centering
\includegraphics[width=\linewidth]{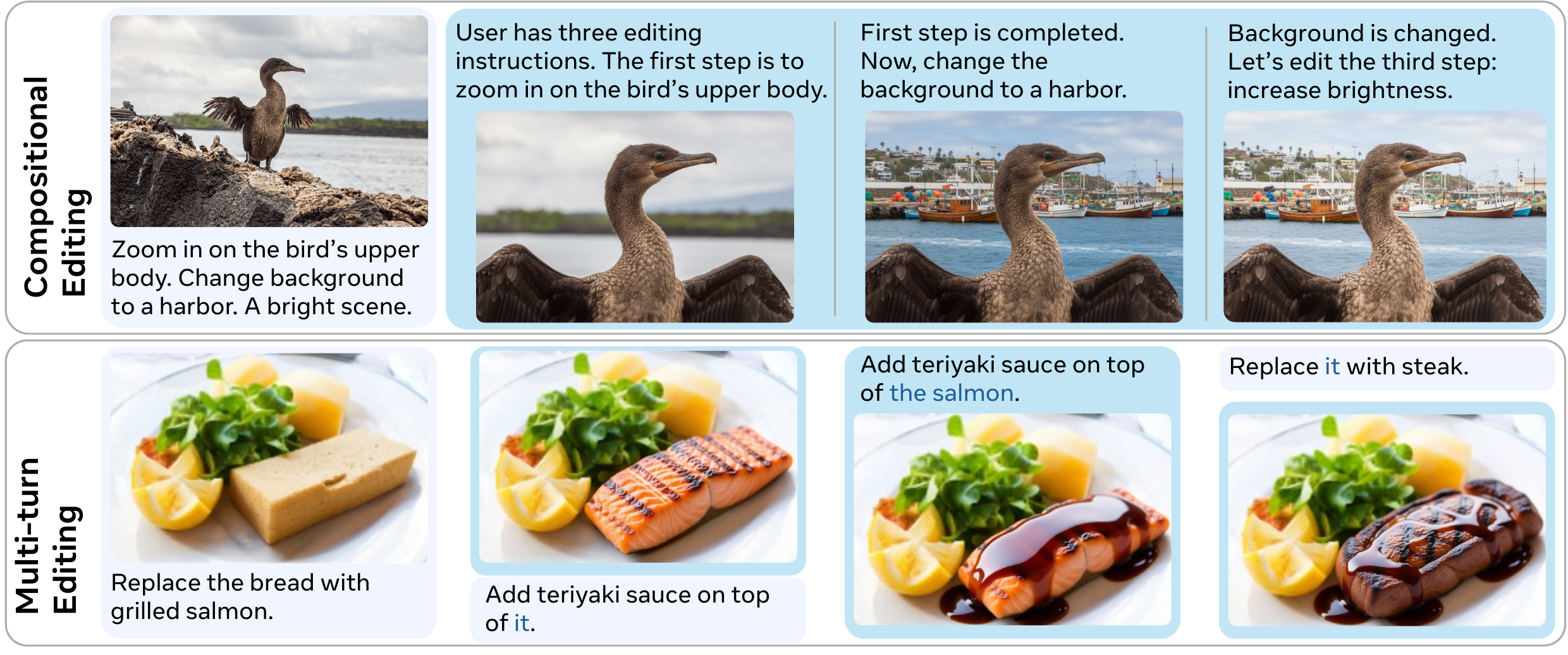}
\includegraphics[width=\linewidth]{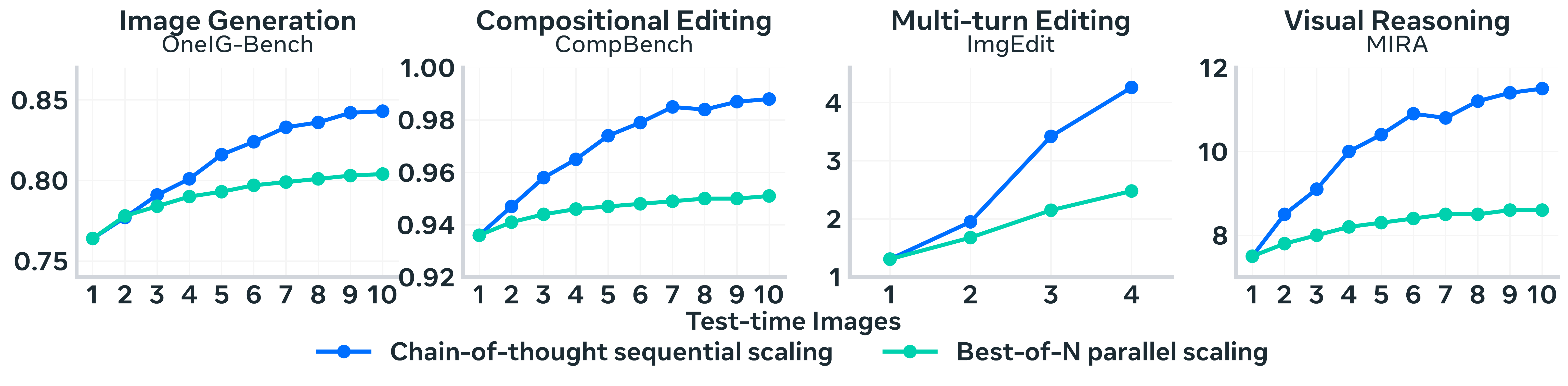}
\caption{
\textbf{Multimodal chain-of-thought enables test-time scaling through emergent cognitive behaviors.} We propose the \textbf{\methodname} framework for unified multimodal models, which induces subgoal decomposition for compositional tasks and unlocks content understanding and memory for multi-turn editing. Controlling the number of test-time images, chain-of-thought sequential scaling outperforms best-of-N parallel scaling across generation and reasoning benchmarks. \teaserlegend
}
\label{fig:teaser}
\end{figure}


\input{sec/1_intro}
\begin{figure}[!t]
    \centering
    \includegraphics[width=0.5\linewidth]{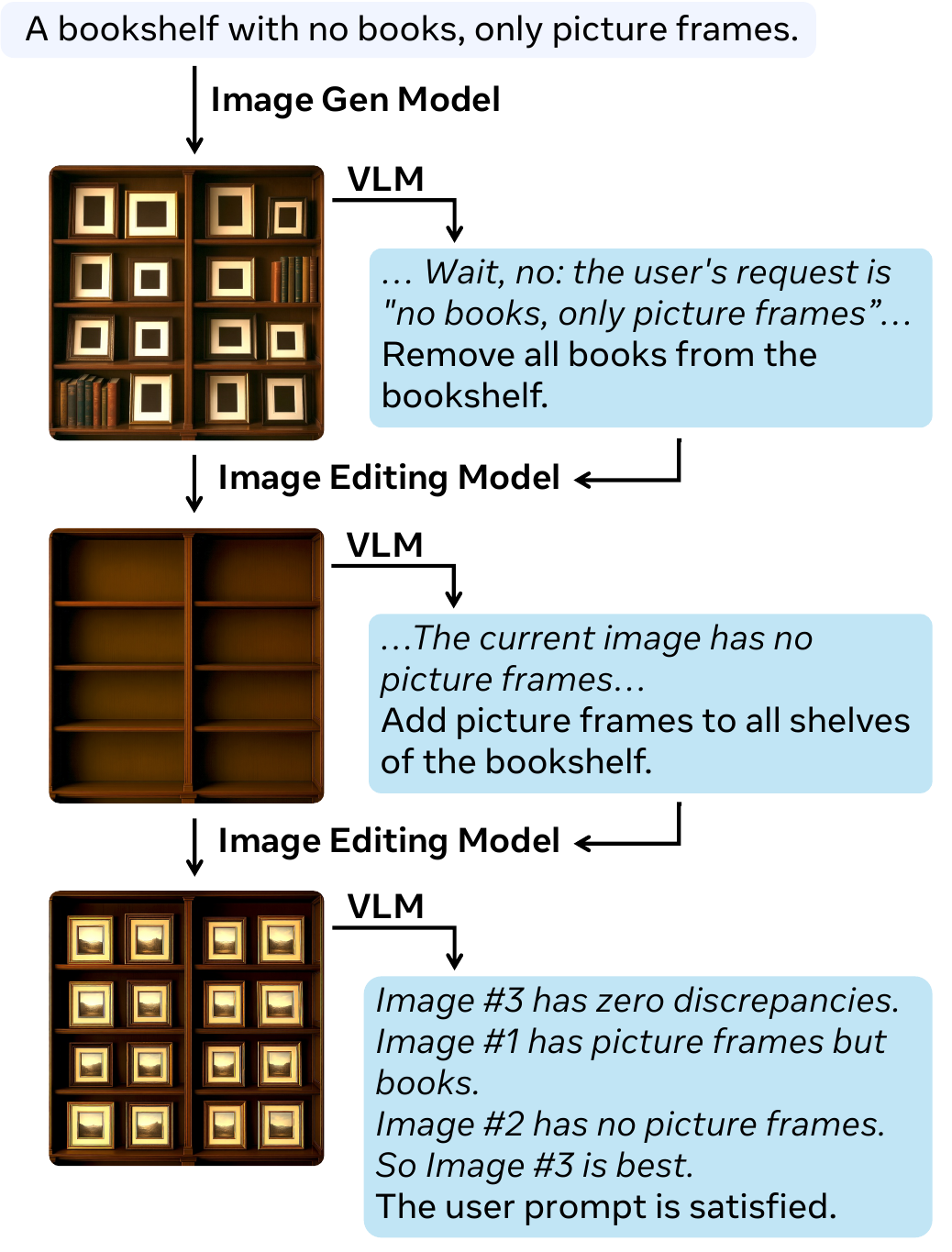}
    \caption{\textbf{Agentic framework for synthesizing chain-of-thought training data.} Starting from a user prompt, an image generation model generates an initial image. A vision-language model then performs verification - evaluating whether the output satisfies the prompt. When unsatisfactory, the VLM engages in explicit subgoal decomposition through thinking tokens, planning concrete improvements, and rewriting editing instructions. This iterative loop continues until verification succeeds, generating multi-turn reasoning trajectories that teach unified models to refine outputs through test-time computation. The explicit reasoning traces of the three models capture how cognitive behaviors emerge from the interplay between generation, verification, and planning. }
    \label{fig:pipeline}
\end{figure}

\input{sec/2_related_works}

\begin{figure}[!t]
    \centering
    \includegraphics[width=\linewidth]{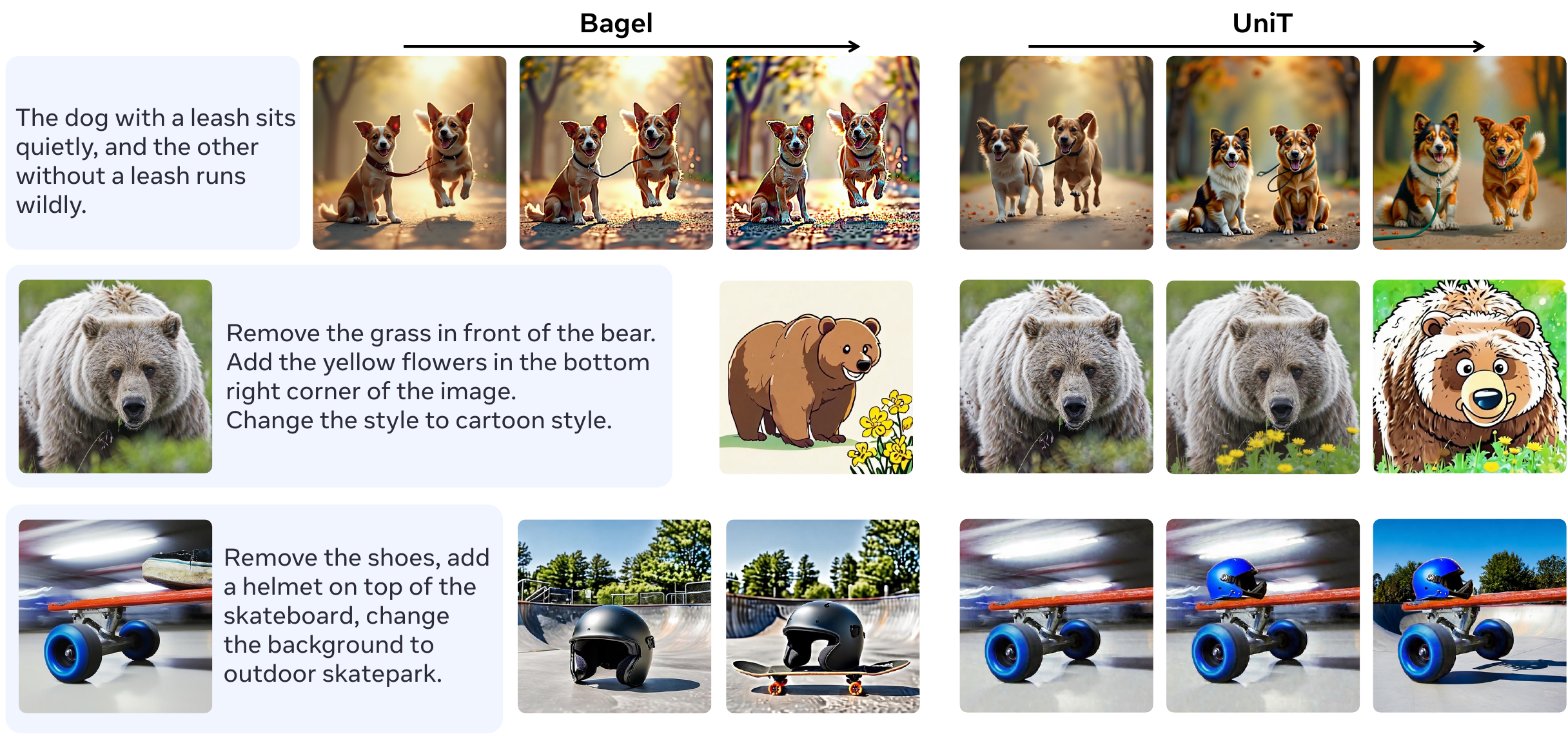}
    \caption{\textbf{UniT enables iterative refinement for compositional instructions through multimodal chain-of-thought reasoning.} UniT exhibits: \textit{(i)} \textbf{error verification and correction}—identifying and fixing constraint violations that Bagel
   misses (top: correcting leash placement and dog action); \textit{(ii)} \textbf{subgoal decomposition with subject consistency}—sequentially addressing instructions while maintaining subject identity across rounds (middle: preserving bear features through style
  transformation, bottom: skateboard consistency); \textit{(iii)} \textbf{quality preservation}—maintaining visual fidelity through iterative refinement rather than degradation (top: reduced artifacts and haloing).
}
    \label{fig:comparison}
\end{figure}

\input{sec/3_method}

\input{sec/4_experiments}

\input{sec/5_analysis}

\input{sec/6_conclusion}

\section*{Acknowledgements}
We thank Hao Tang, Mark Endo, Yuhui Zhang, and Alejandro Lozano for their helpful discussions.

{
    \small
    \bibliographystyle{plainnat}
    \bibliography{paper}
}

\clearpage
\appendix

\input{sec/7_supp}

\end{document}

%% file: preamble_shared.tex
\usepackage{amsmath}
\usepackage{amsfonts}
\usepackage{amssymb}

\makeatletter
\@ifpackageloaded{graphicx}{}{\usepackage{graphicx}}
\@ifpackageloaded{booktabs}{}{\usepackage{booktabs}}
\@ifpackageloaded{multirow}{}{\usepackage{multirow}}
\@ifpackageloaded{xcolor}{}{\usepackage{xcolor}}
\@ifpackageloaded{caption}{}{\usepackage{caption}}  
\makeatother
\usepackage{makecell}
\usepackage{tcolorbox}

\usepackage[linesnumbered,ruled,vlined]{algorithm2e}

\providecommand{\methodname}{}

\newcommand{\budget}[1]{C{\!=\!}#1}

\definecolor{userpromptcolor}{HTML}{F2F5FE}
\definecolor{modeloutputcolor}{HTML}{D1EDFA}

\newcommand{\teaserlegend}{%
  \colorbox{userpromptcolor}{~User input~}
  \colorbox{modeloutputcolor}{~Model output~}%
}

%% file: sec/1_intro.tex
\section{Introduction}
\label{sec:intro}

Unified multimodal models~\citep{deng2025emerging,wu2025janus,zhou2024transfusion} aim to merge vision, language, and more modalities into a single architecture capable of both understanding and generation. Unlike modular pipelines where separate models handle perception, verification, and generation, unified models handle all modalities within one coherent conversation, enabling richer cross-modal grounding, continuous contextual tracking, and seamless interleaving of understanding and generation. However, despite this potential, existing unified models still operate mostly in a \emph{single-pass} mode: they produce an output once, without explicit mechanisms for evaluating, reflecting on, or refining their predictions. This limitation becomes fundamental for tasks that intrinsically require multi-step reasoning and self-correction, such as compositional generation, multi-turn editing, and complex visual reasoning - settings where both humans and AI naturally benefit from extended reasoning.

Recent advances in language models have demonstrated that test-time scaling (TTS)—allocating additional computational resources during inference through extended chain-of-thought reasoning, verification, and iterative refinement~\citep{jaech2024openai,guo2025deepseek,snell2024scaling}—enables substantial performance gains on complex reasoning tasks in mathematics~\citep{cobbe2021training}, coding~\citep{chen2021evaluating}, and logic~\citep{srivastava2023beyond}. Early work on multimodal chain-of-thought has shown similar benefits for single-round visual understanding and generation~\citep{visualcot,fang2025got,xiao2025mindomni,huang2025interleaving,chern2025thinking}. Yet, test-time scaling for unified multimodal models remains largely unexplored. The challenge is nontrivial: test-time scaling requires capabilities that currently scatter across specialized models (image generation models for generation, vision-language models for verification, image editing models for refinement). Bridging this gap requires a unified framework that systematically integrates data synthesis, model training, and inference mechanisms for multimodal test-time scaling. This motivates the central question:

\textbf{\textit{How to enable scalable multimodal inference that allows unified models to iteratively generate, reflect, and refine?}}

We introduce \textbf{UniT}, a unified framework for multimodal chain-of-thought test-time scaling. Scalable multimodal inference requires the tight integration of three components: \textbf{\textit{(i) Agentic data synthesis}} to induce cognitive behaviors through multi-round trajectories. We develop an automated pipeline (Fig.~\ref{fig:pipeline}) where vision-language models iteratively critique and image editing models refine generated images with explicit chain-of-thought reasoning. This naturally produces training data exhibiting three critical cognitive behaviors \citep{gandhi2025cognitive}: \textbf{verification}—evaluating outputs against instructions; \textbf{subgoal decomposition}—breaking complex instructions into sequential planning steps; \textbf{content memory}—maintaining understanding of visual content across rounds through unified multimodal context. \textbf{\textit{(ii) Unified model training}} to enable the model to internalize multimodal reasoning patterns. We collect approximately 12K multi-round trajectories and fine-tune the Bagel unified multimodal model~\citep{deng2025emerging} for 700 H100 hours, enabling it to perform both understanding and refinement without switching models. \textbf{\textit{(iii) Multimodal test-time scaling}} at inference with flexible computational budget. The trained model performs all reasoning, generation, and refinement iteratively through explicit multimodal chain-of-thought thinking, allocating more rounds to more challenging tasks.

The synergy of these components enables the model to act as a single, coherent multimodal reasoner capable of self-evaluation and iterative improvement. The UniT framework exhibits strong test-time scaling behavior (Fig.~\ref{fig:teaser}) with emergent capabilities. Most notably, \textit{models trained on shorter reasoning trajectories (averaging 3.6 rounds) effectively generalize to longer inference chains at test time (averaging 4.7 rounds)} (Fig.~\ref{fig:round_distribution}), echoing patterns previously seen only in text-only models~\citep{snell2024scaling}. Furthermore, chain-of-thought sequential scaling substantially outperforms best-of-N parallel sampling, achieving comparable performance with 2.5$\times$ less computational cost (Fig.~\ref{fig:teaser}). This demonstrates that iterative refinement with explicit reasoning provides more efficient use of inference compute than parallel sampling. Critically, UniT achieves 5.56\% improvement on CompBench multi-object editing, 2.95 human preference scores on ImgEdit multi-turn editing, and 10.34\% on OneIG instruction following compared to single-pass generation. Moreover, it improves out-of-distribution visual reasoning on MIRA by 53.33\%, \textit{establishing multimodal chain-of-thought test-time scaling as a unified paradigm that benefits both generation and comprehension}.

We summarize our contributions as follows:
\begin{itemize}
    \item \textbf{Unified multimodal test-time scaling.} We propose \textbf{UniT}, a unified framework for multimodal chain-of-thought test-time scaling, integrating agentic data synthesis, unified model training, and test-time scaling mechanisms.
    \item \textbf{Emergent extrapolation to longer reasoning chains.} We demonstrate that models trained on shorter trajectories generalize to longer inference chains at test time, extrapolating beyond the training distribution.
    \item \textbf{Broad improvements across multimodal tasks.} UniT achieves substantial gains on compositional generation/editing, multi-turn editing, and visual reasoning, establishing chain-of-thought test-time scaling as a unified paradigm for both generation and understanding tasks.
\end{itemize}

%% file: sec/2_related_works.tex
\section{Related Works}
\label{sec:related_works}

\noindent\textbf{Test-time scaling.} Test-time scaling allocates additional computation during inference to improve model performance. We distinguish two primary approaches: parallel and sequential methods. Parallel scaling generates multiple independent candidates and selects the best via criteria such as best-of-N sampling~\citep{brown2024largelanguagemonkeysscaling, levi2024simplemodelinferencescaling} or majority voting~\citep{irvine2023rewardingchatbotsrealworldengagement}, typically guided by outcome reward models~\citep{xin2024deepseekproveradvancingtheoremproving, ankner2024critiqueoutloudrewardmodels}. Sequential scaling~\citep{snell2024scalingllmtesttimecompute, hou2025advancinglanguagemodelreasoning, lee2025evolvingdeeperllmthinking} enables iterative refinement where models critique and improve outputs across multiple rounds. Self-refinement approaches~\citep{madaan2024self} exemplify this by having models explicitly reason about deficiencies and produce progressively better solutions. Tree-based methods such as Monte Carlo Tree Search~\citep{liu2024dontthrowawayvalue, zhang2023planninglargelanguagemodels, zhou2024languageagenttreesearch, choi2023kcts} and \textsc{REBASE}~\citep{wu2024inference} occupy a middle ground, using process reward models~\citep{lightman2023letsverifystepstep, wang-etal-2024-math, wang2024helpsteer2opensourcedatasettraining} to guide structured search. Recent breakthroughs including o1~\citep{jaech2024openai} and DeepSeek-R1~\citep{guo2025deepseek} demonstrate that reinforcement learning enables effective utilization of extended inference computation. Budget forcing~\citep{snell2024scalingllmtesttimecompute, tyen2024lamini} trains models to produce reasoning chains of controllable cost by varying computational budgets during training. While most test-time scaling research focuses on text-only reasoning, recent work has explored search-based methods for image and video generation~\citep{he2025scaling, liu2025video}. However, unified multimodal test-time scaling interleaving text and images remains largely unexplored, which we address in this work.

\noindent\textbf{Unified multimodal models.} Unified models that jointly handle understanding and generation have attracted substantial interest. Autoregressive approaches~\citep{wu2025janus, januspro2025, lu2024unified, qu2024tokenflow, chameleon, emu3} extend next-token prediction to both text and discrete image tokens. Additional diffusion methods~\citep{dream-llm, wu2024next, pan2025transfer, tong2024metamorph} augment language models with external diffusion modules for image generation. Unified integrated transformers~\citep{deng2025emerging, liang2024mixture, janusflow2024, shi2024llamafusion, transfusion} deeply integrate language modeling and diffusion within single architectures. Our work builds upon Bagel~\citep{deng2025emerging}, pretrained on large-scale interleaved text-image sequences. Our framework generalizes to all three paradigms as they naturally handle interleaved multimodal inputs and outputs.

\noindent\textbf{Multimodal chain-of-thought.} Chain-of-thought reasoning~\citep{wei2022chain} has proven effective for text-based problem-solving, motivating extensions to multimodal tasks. Visual chain-of-thought methods~\citep{visualcot, chainofimage, visualsketchpad, mvot, visualizationofthought, huang2025vchain} incorporate visual representations into reasoning steps for multimodal understanding. Recent work explores interleaved reasoning~\citep{huang2025interleaving, gu2025thinkmorph} across text and visual modalities. Uni-CoT~\citep{qin2025uni} further demonstrates a unified model that couples macro- and micro-level reasoning for vision-language understanding, but it does not study compute scaling or iterative editing. In text-to-image generation, studies investigate whether explicit reasoning improves generation quality~\citep{fang2025got, xiao2025mindomni, deng2025emerging, jiang2025t2i, gu2025improving}. Reflection-based approaches~\citep{zhuo2025reflection, wu2025omnigen2, chern2025thinking} iteratively critique and refine generated images. Our work differs by focusing on both semantic correctness and visual quality through test-time scaled refinement, while demonstrating that multimodal chain-of-thought benefits both generation and understanding tasks as a unified paradigm.

%% file: sec/3_method.tex
\section{Method}
\label{sec:method}

We extend test-time compute scaling from text-only reasoning to unified multimodal models. As illustrated in Fig.~\ref{fig:pipeline}, we develop an agentic framework to automatically collect chain-of-thought training data, then control test-time computational budget through iterative refinement rounds. Sequential chain-of-thought scaling outperforms parallel best-of-N sampling (Fig.~\ref{fig:teaser}) while inducing emergent cognitive behaviors—verification and subgoal decomposition.

\noindent \textbf{Key distinction.} The multi-model agentic framework described in Sec.~\ref{sec:data_collection} is used \textit{solely for synthesizing training data}. At inference time (Sec.~\ref{sec:budget_forcing}), we use only the single unified BAGEL model~\citep{deng2025emerging}, which performs all planning, generation, reflection, and refinement operations without external models.

\subsection{Multimodal Chain-of-Thought Data}
\label{sec:data_collection}

\noindent \textbf{Agentic data collection pipeline.} Our automated pipeline synthesizes multimodal chain-of-thought trajectories through iterative reflection-editing (Fig.~\ref{fig:pipeline}):

\begin{enumerate}
    \item \textbf{Prompt generation}: Llama-4-Scout-17B-16E \citep{meta2025llama4scout} generates 20K diverse prompts covering compositional attributes, spatial relations, and complex multimodal tasks based on \citep{T2I-CoReBench}.
    \item \textbf{Initial generation}: Flux Pro \citep{flux} produces initial images from prompts. For complex prompts, the VLM (Qwen3-VL) decomposes the prompts into subgoals and executes the first step in initial generation.
    \item \textbf{Reflection}: Qwen3-VL \citep{qwen2025vl} evaluates whether the image satisfies the prompt. If not, it generates explicit chain-of-thought reasoning, identifying deficiencies, planning improvements, and specifying editing instructions.
    \item \textbf{Refinement}: Flux Kontext \citep{labs2025flux} or Qwen-Image-Edit \citep{wu2025qwen} applies editing instructions.
    \item \textbf{Iteration}: Steps 3-4 repeat until the VLM determines the output satisfies the prompt.
\end{enumerate}

\noindent \textbf{Cognitive behaviors.} As demonstrated in Fig.~\ref{fig:teaser}, \ref{fig:pipeline}, this agentic pipeline naturally induces three critical cognitive behaviors: \textbf{(i) verification}—VLMs evaluate outputs against specifications to determine when refinement is needed; \textbf{(ii) subgoal decomposition}—complex compositional tasks are planned in sequential editing steps; \textbf{(iii) content memory}—the model maintains understanding of image content across refinement rounds through unified multimodal context.

\noindent \textbf{Data filtering.}
We apply quality filtering to ensure training efficiency:

\begin{itemize}
    \item \textbf{Length constraint}: Trajectories exceeding 8 rounds are removed to balance efficiency with reasoning depth.

    \item \textbf{Quality regression}: Trajectories where the final image has worse instruction-following quality than any of the first three images (measured by Qwen3-VL) are removed.

    \item \textbf{Relevance filtering}: Individual rounds with editing prompts semantically irrelevant to the original task (measured by Llama-4-Scout-17B-16E) are removed.

    \item \textbf{Minimal visual changes}: Rounds with perceptual editing distance below LPIPS $< 0.03$~\citep{zhang2018perceptual} between consecutive images are removed.

    \item \textbf{Benchmark deduplication}: Training prompts are deduplicated from evaluation benchmarks using 5-gram matching to prevent data leakage.
\end{itemize}

This filtering retains 12k high-quality trajectories for training.

\input{tables/oneig}

\subsection{Training and Inference}
\label{sec:training_inference}

\noindent \textbf{Training.}  We use Bagel \citep{deng2025emerging}, a unified multimodal architecture with understanding and generation capabilities, trained on the dataset from Sec.~\ref{sec:data_collection} for 700 H100 hours. To simulate user prompts for multi-turn editing, 10\% of intermediate image editing instructions don't require losses. 

\noindent \textbf{Inference.}  We adopt a framework that incorporates two complementary classifier-free guidance (CFG) schemes applied in a nested manner: (1) text CFG, conditioning with versus without the current text instruction, and (2) image CFG, conditioning
  with versus without all images in the generation history (including both the initial input image if exists and previously generated outputs). Formally, let $v_t$ denote the fully conditional prediction, $v_{t,\text{unc}}$ the text-unconditional prediction, and $v_{i,\text{unc}}$ the image-unconditional prediction. We apply CFG sequentially: first text guidance $v_{\text{text}} = v_{t,\text{unc}} + s_t(v_t - v_{t,\text{unc}})$, then image guidance $v_{\text{final}} = v_{i,\text{unc}} + s_i(v_{\text{text}} - v_{i,\text{unc}})$, with scales $s_t{=}4.0$ and $s_i{=}2.0$. The nested application—where image guidance is applied on top of the text-guided
  prediction—enables independent control over prompt adherence and visual consistency. This strategy helps maintain strong alignment with text instructions while preserving structural coherence across
  multi-turn editing sequences, with notable benefits for generation quality and consistency in iterative refinement workflows.

The original untrained Bagel model can also be forced to reason chain-of-thought with our inference code. However, the image quality degrades quickly as context images scale and the Bagel model can't verify its image outputs properly. It usually hallucinates visual content according to user prompts. Thus the untrained Bagel model is not feasible for practical multimodal chain-of-thought reasoning. Training is required to enable effective multimodal chain-of-thought reasoning.

\subsection{Budget Forcing for Test-time Scaling}
\label{sec:budget_forcing}

We adapt budget forcing~\citep{tyen2024lamini} from text-only to multimodal test-time scaling. While text-based methods control reasoning tokens, we control image generation rounds, which dominate inference latency. At test time, the unified BAGEL model performs all operations autonomously—planning, generation, reflection, and refinement—without relying on external models.

\noindent \textbf{Controlling computational budget.} At inference, we specify computational budget $C$ as the number of image generation rounds. Each round consists of textual chain-of-thought reasoning followed by image generation or editing. To enforce budget $C$:

\begin{itemize}
    \item \textbf{Forcing extended reasoning}: If the model terminates before $C$ rounds, we suppress EOS, append ``\textit{Let's edit the image}", wait for reasoning completion, then force image generation.

    \item \textbf{Budget constraint}: If the model generates more than $C$ images, we use only the final image from round $C$.
\end{itemize}

This enables studying sequential chain-of-thought scaling (iterative refinement building on previous outputs) versus parallel best-of-N scaling (generating $N$ independent images and selecting the best). We present detailed scaling analysis in Sec.~\ref{subsec:sequential_vs_parallel}.

\begin{figure}[!t]
    \centering
    \includegraphics[width=0.6\linewidth]{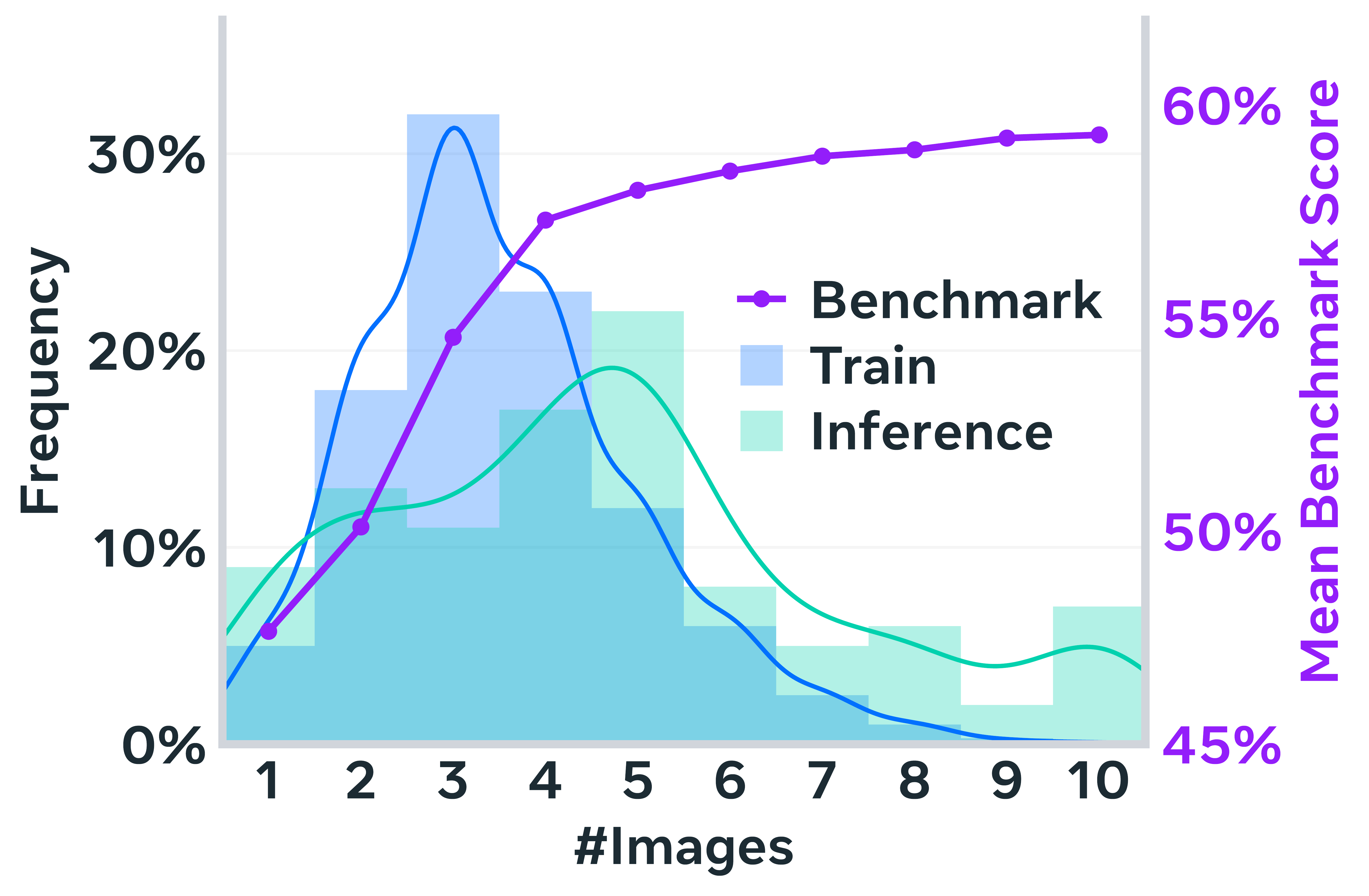}
    \caption{\textbf{Training vs. inference round distribution demonstrates beyond-training generalization.} The model is trained on trajectories averaging 3.6 refinement rounds, but effectively generalizes to longer inference chains averaging 4.7 rounds at test time. This distribution shift reveals the model's emergent ability to extend inference beyond its training distribution, a key property of effective test-time scaling.}
    \label{fig:round_distribution}
\end{figure}

\noindent \textbf{Beyond-training generalization.} Models trained on trajectories averaging 3.6 rounds generalize to longer inference chains averaging 4.7 rounds at test time (Fig.~\ref{fig:round_distribution}), exhibiting problem-solving capabilities beyond their training distribution. This establishes test-time compute as a general paradigm for unified multimodal models.

%% file: tables/oneig.tex
\begin{table}[!t]
\centering
\small
\resizebox{0.9\linewidth}{!}{%
\begin{tabular}{l|cc|ccc|c}
\toprule
\multirow{2}{*}{\textbf{Method}} & \multicolumn{6}{c}{\textbf{Alignment}$\uparrow$}  \\
\cline{2-7}
& \textbf{NP} & \textbf{T\&P} & \textbf{Short} & \textbf{Medium} & \textbf{Long} & \textbf{Overall} \\
\midrule
Janus-Pro~\citep{chen2025janus} & 0.557  & 0.533  & 0.609   & 0.548 & 0.515 & 0.552 \\
BLIP3-o~\citep{chen2025blip3} & 0.719  & 0.671  & 0.754   & 0.712 & 0.674 & 0.706 \\
Bagel~\citep{deng2025bagel} & 0.776  & 0.734  & 0.782   & 0.769 & 0.759 & 0.764 \\
Bagel+CoT~\citep{deng2025bagel} & 0.798 & 0.767 & 0.824 & 0.793 & 0.767 & 0.790 \\
\midrule
\methodname & \textbf{0.853} & \textbf{0.844} & \textbf{0.859} & \textbf{0.849} & \textbf{0.812} & \textbf{0.843} \\
\bottomrule
\end{tabular}%
 }
\caption{\textbf{Compositional generation, OneIG-Bench}. NP denotes the natural
language prompt. T\&P denotes the tag-based and phrase-based prompt. Short, Medium and Long represent the length of the prompts, where Short denotes the number of words is less than 30, Medium denotes the number between 30 and 60, and Long denotes the number exceeding 60. Bagel+CoT indicates Bagel with text-only chain-of-thought.
}
\label{tab:oneig_alignment}
\end{table}

%% file: sec/4_experiments.tex
\section{Experiments}
\label{sec:experiments}

We evaluate on text-to-image generation, compositional editing, multi-turn editing, and visual reasoning benchmarks. Chain-of-thought TTS achieves substantial gains across both generation and understanding tasks.

\subsection{Experiment Settings}
\label{subsec:exp_settings}

\noindent \textbf{Evaluation setup.} We evaluate with computational budgets $\budget{1}$ to $\budget{10}$ (maximum due to GPU memory), controlled via Sec.~\ref{sec:budget_forcing}. For ImgEdit multi-turn editing, we apply $C$ refinement rounds independently to each of three sequential turns, with maximum $\budget{4}$ per turn due to memory constraints.

\noindent \textbf{Baselines.} We compare: (1) \textbf{Bagel}, base model without chain-of-thought; (2) \textbf{Bagel+CoT}, textual thinking only; (3) \methodname, our full multimodal chain-of-thought with interleaved text and image reasoning. We report $\budget{10}$ rounds ($\budget{4}$ for ImgEdit) unless specified otherwise.

\noindent \textbf{Human evaluation protocol.} For ImgEdit multi-turn editing, 3 expert annotators with computer vision backgrounds independently rate outputs on a 0-10 scale across three criteria: content memory (tracking edits across turns), content understanding (correctly interpreting instructions), and version backtracking (maintaining coherence). We evaluate 100 randomly sampled test examples per method. Inter-annotator agreement is high (Krippendorff's $\alpha=0.82$), and final scores are averaged across annotators.

\subsection{Compositional Generation and Editing}
\label{subsec:compositional}
We evaluate on OneIG-Bench-EN~\citep{chang2025oneig} for compositional generation (Table~\ref{tab:oneig_alignment}), achieving 10.34\% improvement over base model at $\budget{10}$. On CompBench~\citep{jia2025compbench} multi-object compositional editing subset (Table~\ref{table:compbench}), we achieve 5.56\% improvement from $\budget{1}$ to $\budget{10}$. Fig.~\ref{fig:teaser} demonstrates that iterative refinement with explicit reasoning enables better compositional understanding and generation.

\input{tables/imgedit}

\subsection{Multi-Turn Editing}
\label{subsec:multiturn}

We evaluate on ImgEdit~\citep{ye2025imgedit} multi-turn editing subset (Table~\ref{tab:multi_turn}), achieving 225.19\% improvement from $\budget{1}$ to $\budget{4}$. This demonstrates that maintained context and reasoning chains are critical for multi-turn interactions. The model's content memory through unified multimodal context enables coherent interactions as computational budget increases (Fig.~\ref{fig:teaser}).

\subsection{Visual Reasoning}
\label{subsec:visual_reasoning}

On MIRA~\citep{zhou2025visualizing} for out-of-distribution visual reasoning (Table~\ref{tab:mira_results}), we achieve 53.33\% improvement from $\budget{1}$ to $\budget{10}$ (Fig.~\ref{fig:teaser}). The remaining gap to frontier models (GPT-5, Qwen2.5-VL-72B) reflects base model capability differences: these models benefit from significantly larger scale and proprietary training data. Our key contribution is methodological, demonstrating that TTS successfully transfers to multimodal domains; as base unified models improve, the UniT framework directly benefits. This establishes multimodal chain-of-thought test-time scaling as a unified paradigm enhancing both generation and comprehension tasks. The cognitive behaviors induced by our agentic framework (Fig.~\ref{fig:pipeline})—verification, subgoal decomposition, and content memory—transfer from generation to understanding tasks.

\subsection{Qualitative Results}
\label{subsec:qualitative}

\input{tables/compbench}

Fig.~\ref{fig:teaser} and Fig.~\ref{fig:comparison} demonstrate how multimodal chain-of-thought test-time scaling progressively refines outputs through iterative reasoning, revealing the three cognitive behaviors—verification, subgoal decomposition, and content memory.

\noindent \textbf{Compositional generation.} For complex compositional prompts, Round 1 produces partial solutions with compositional errors—missing objects, incorrect attributes, or violated spatial constraints. The model's explicit reasoning identifies deficiencies, and through subgoal decomposition, breaks corrections into sequential steps. By Round 10, the output achieves precise alignment with all requirements, demonstrating systematic error correction through explicit reasoning.

\noindent \textbf{Visual reasoning.} For MIRA geometry tasks, the model's chain-of-thought reveals how verification supports iterative problem-solving. Early rounds may produce incorrect analyses, but the reasoning shows self-critique, allowing identification of flaws and revision in subsequent rounds. Subgoal decomposition breaks complex reasoning into steps. These examples demonstrate that cognitive behaviors transfer from generation to understanding tasks.

%% file: tables/imgedit.tex
\begin{table}[!t]
\centering
\small
\resizebox{0.9\linewidth}{!}{%
\begin{tabular}{l|ccc|c}
\toprule
& \textbf{Content} & \textbf{Content} & \textbf{Version} & \\
\textbf{Method} & \textbf{Memory} & \textbf{Understand} & \textbf{Backtrack} & \textbf{Overall} \\
\midrule
Bagel \citep{deng2025bagel} & 1.76 & 1.34 & 0.82 & 1.31 \\
Bagel+CoT \citep{deng2025bagel} & 2.24 & 2.67 & 0.84 & 1.92 \\
\midrule
\methodname & \textbf{4.29} & \textbf{5.02}  & \textbf{3.48} & \textbf{4.26} \\
\bottomrule
\end{tabular}
 }
\caption{\textbf{Multi-turn editing, ImgEdit}. Human evaluation score from 0-10, normalized over 3 turns of editing.}
\label{tab:multi_turn}
\end{table}

%% file: tables/compbench.tex
\begin{table}[!t]
\centering
\small
\setlength{\tabcolsep}{3pt}
\resizebox{0.8\linewidth}{!}{%
\begin{tabular}{l|cc|ccc|c}
\toprule
& \multicolumn{2}{c|}{\textbf{Foreground}} & \multicolumn{3}{c|}{\textbf{Background}} & \\
\textbf{Method} & LC-T$\uparrow$ & LC-I$\uparrow$ & PSNR(dB)$\uparrow$ & SSIM$\uparrow$ & LPIPS$\downarrow$ & \textbf{Overall}$\uparrow$ \\
\midrule
HQ-Edit \citep{hui2024hqedit} & 19.163 & 0.757 & 12.987 & 0.412 & 0.421 & 0.007 \\
UltraEdit \citep{zhao2024ultraedit} & 20.022 & 0.795 & 22.326 & 0.719 & 0.164 & 0.609 \\
AnyEdit \citep{yu2024anyedit} & 19.875 & 0.809 & 22.789 & 0.697 & 0.129 & 0.622 \\
SEED-X \citep{ge2024seedx} & 19.092 & 0.795 & 20.638 & 0.788 & 0.138 & 0.595 \\
GoT \citep{fang2025got} & 19.919 & 0.804 & 21.296 & 0.826 & 0.127 & 0.668 \\
Step1X-Edit \citep{liu2025step1x} & 20.213 & 0.828 & 22.696 & 0.873 & 0.089 & 0.844 \\
FLUX.1 Kontext \citep{labs2025flux} & 20.983 & 0.836 & 24.013 & 0.938 & 0.064 & 0.965 \\
Qwen-Image-Edit \citep{wu2025qwen} & 21.058 & 0.836 & 21.927 & 0.810 & 0.121 & 0.782 \\
Bagel \citep{deng2025bagel} & 20.434 & 0.842 & 24.370 & 0.917 & 0.069 & 0.936 \\
Bagel+CoT \citep{deng2025bagel} & 20.658 & 0.846 & 24.691 & 0.926 & 0.065 & 0.956 \\
\midrule
\methodname & \textbf{21.127} & \textbf{0.854} & \textbf{25.442} & \textbf{0.942} & \textbf{0.055} & \textbf{0.988} \\
\bottomrule
\end{tabular}%
 }
\caption{\textbf{Multi-object editing, CompBench.} LC-T denotes local CLIP scores between the edited foreground and the local description. LC-I refers to the CLIP image similarity between the foreground edited result and ground truth (GT) image. Overall scores are computed using min-max normalization for each metric.}
\label{table:compbench}
\end{table}

%% file: sec/5_analysis.tex
\section{Discussion}
\label{sec:discussion}

\begin{figure}[!t]
    \centering
    \includegraphics[width=0.7\linewidth]{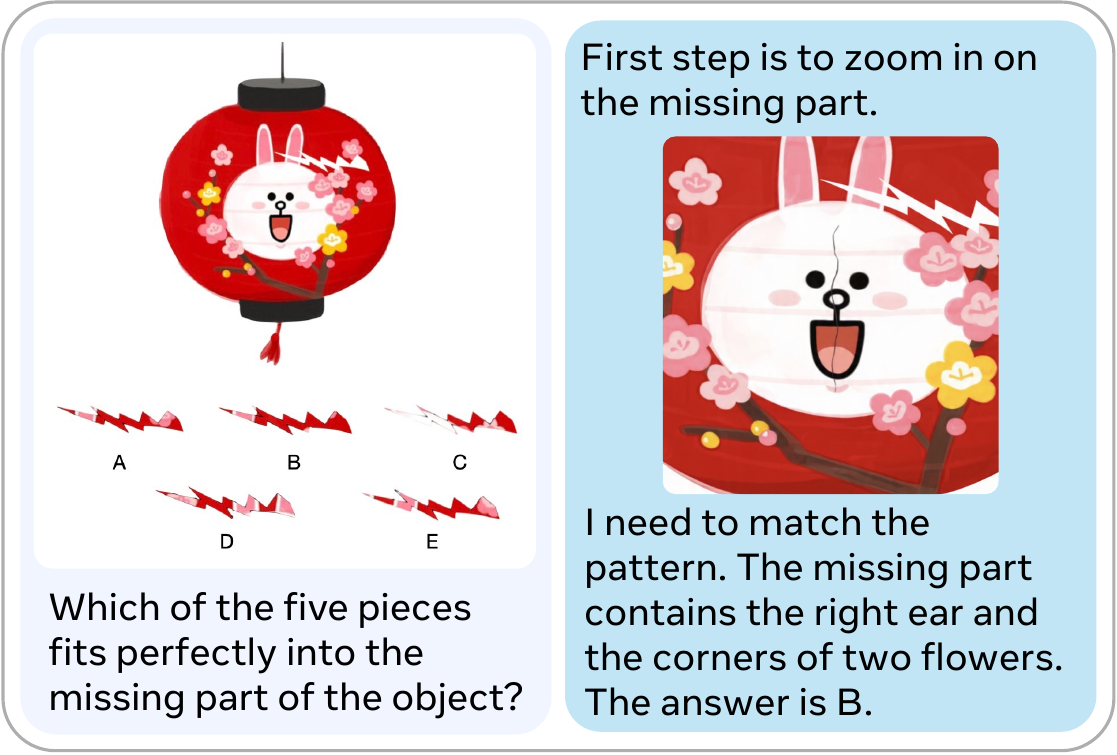}
    \caption{\textbf{Chain-of-thought visual reasoning on MIRA.} The model decomposes the puzzle into subgoals (zoom in, identify patterns) before selecting the matching piece, demonstrating cognitive behaviors transferring from generation to understanding tasks.}
    \label{fig:reasoning}
\end{figure}

We analyze factors contributing to effective multimodal test-time scaling: sequential versus parallel scaling, cognitive behaviors, and data quality.

\noindent \textbf{Experimental protocol.} Unless specified otherwise, we use computational budget $\budget{10}$ ($\budget{4}$ for ImgEdit) and report: alignment score for OneIG-Bench, overall normalized score for CompBench, human evaluation score (0-10) for ImgEdit, and accuracy for MIRA.

\subsection{Sequential vs. Parallel Scaling}
\label{subsec:sequential_vs_parallel}

We compare chain-of-thought sequential scaling against best-of-N parallel scaling. Following~\cite{tyen2024lamini}, sequential scaling builds on intermediate results through iterative refinement, while parallel scaling generates outputs independently.

\noindent \textbf{Setup.} For both approaches, we control the number of generated images ($C{=}N$) from 1 to 10. Sequential scaling uses budget forcing (Sec.~\ref{sec:budget_forcing}). Parallel scaling generates $N$ independent images and selects the best via HPSv3~\citep{ma2025hpsv3}.

\noindent \textbf{Compute accounting.} We use the number of generated images as our compute metric ($C{=}N$), which accurately reflects computational cost because: (i) image generation via diffusion models dominates wall-clock time—as noted in Sec.~\ref{sec:budget_forcing}, image generation rounds control inference latency; (ii) marginal text tokens from VLM reflection have negligible effect on latency compared to diffusion sampling; (iii) computational cost scales linearly with the number of images generated. We exclude selection costs (HPSv3 for parallel scaling, VLM verification for sequential scaling) from our comparison as these represent arbitrary model choices rather than fundamental algorithmic requirements. Under this metric, sequential scaling achieves comparable performance to parallel with 2.5× fewer generated images (e.g., $\budget{4}$ sequential matches $N{=}10$ parallel on OneIG-Bench).

\noindent \textbf{Results.} Sequential scaling consistently outperforms parallel scaling across all tasks (Fig.~\ref{fig:teaser}). At $\budget{10}$ ($\budget{4}$ for ImgEdit), sequential achieves 4.85\% improvement over parallel on OneIG-Bench, 3.89\% on CompBench, 71.77\% on ImgEdit, and 33.72\% on MIRA. This advantage manifests through:

\begin{itemize}
    \item \textbf{Steeper scaling slopes}: Sequential achieves larger performance improvements per additional image, indicating more efficient test-time compute use.

    \item \textbf{Sustained improvements}: Sequential shows continued gains up to $\budget{10}$, while parallel plateaus after a few samples.

\end{itemize}

Iterative refinement with explicit chain-of-thought reasoning provides more effective test-time compute scaling than independent sampling. The advantage stems from sequential scaling accumulating successful edits and learning from previous iterations: each round builds on prior images with explicit CoT corrections, leveraging expanded textual context (reflections, plans). Parallel scaling generates independent samples without inter-sample learning, explaining why it plateaus earlier—it cannot systematically refine toward the target.

\noindent \textbf{Latency considerations.} Sequential and parallel scaling serve complementary use cases: sequential optimizes performance while parallel optimizes latency. Sequential scaling also benefits from unique acceleration techniques including speculative decoding, KV-cache reuse across rounds, and early stopping when the model determines satisfaction, which can significantly reduce the latency gap in practice.

\noindent \textbf{Unified model vs. modular pipeline.} At inference, UniT performs all reasoning, generation, and refinement within a single model, unlike the multi-model agentic pipeline used for data synthesis (Sec.~\ref{sec:data_collection}). While the teacher pipeline (Flux Pro + Qwen3-VL) scores slightly higher due to frontier-scale components, UniT offers faster inference by eliminating inter-model communication overhead, seamless multimodal context within a single architecture, and practical single-model deployability.

\input{tables/mira}

\subsection{Ablation on Cognitive Behaviors}
\label{subsec:cognitive_ablation}

\noindent \textbf{Setup.} We train three ablated models, each removing one cognitive behavior: (1) \textbf{w/o Verification}, reflection stages do not evaluate quality; (2) \textbf{w/o Subgoal Decomposition}, planning stages removed; (3) \textbf{w/o Content Memory}, multimodal context not maintained across rounds.

\noindent \textbf{Results.} Table~\ref{tab:cognitive_ablation} shows task-specific sensitivities. Removing subgoal decomposition particularly hurts compositional tasks (3.8\% and 2.5\% drops on OneIG-Bench and CompBench), demonstrating its importance for planning multi-step generations. Most critically, removing content memory devastates multi-turn editing with 42.5\% relative drop on ImgEdit (from 4.26 to 2.45), while minimally impacting single-turn tasks (1.0-1.5\% drops). Removing verification most significantly impacts visual reasoning (1.9\% drop on MIRA), highlighting its importance for validating reasoning steps. These patterns confirm each cognitive behavior serves a distinct functional role.

\begin{table}[t]
\centering
\small
\resizebox{0.9\linewidth}{!}{%
\begin{tabular}{lcccc}
\toprule
\textbf{Configuration} & \textbf{OneIG} & \textbf{CompBench} & \textbf{ImgEdit} & \textbf{MIRA} \\
& \textbf{Align (\%)} & \textbf{Overall (\%)} & \textbf{Score} & \textbf{Acc (\%)} \\
\midrule
All behaviours & 84.3 & 98.8 & 4.26 & 11.5 \\
w/o Verification & 81.2 (-3.1) & 96.8 (-2.0) & 3.55 (-0.71) & 9.6 (-1.9) \\
w/o Subgoal Decomp.& 80.5 (-3.8) & 96.3 (-2.5) & 3.75 (-0.51) & 10.3 (-1.2) \\
w/o Content Memory & 82.8 (-1.5) & 97.8 (-1.0) & 2.45 (-1.81) & 10.8 (-0.7) \\
\bottomrule
\end{tabular}%
}
\caption{\textbf{Cognitive behavior ablation.} Impact of removing verification, subgoal decomposition, or content memory from our agentic framework.}
\label{tab:cognitive_ablation}
\end{table}

\subsection{Data Quality Analysis}
\label{subsec:data_quality}

We analyze the impact of data quality by ablating individual filters from our curation pipeline (Sec.~\ref{sec:data_collection}).

\noindent \textbf{Setup.} We train three ablated models, each removing one quality filter: (1) \textbf{w/o Quality regression filter}, including trajectories where refinement degrades quality; (2) \textbf{w/o Relevance filtering}, including rounds with semantically irrelevant editing prompts; (3) \textbf{w/o Minimal visual changes filter}, including rounds with negligible visual refinement.

\noindent \textbf{Results.} Table~\ref{tab:data_quality} shows task-specific sensitivities. Removing relevance filtering causes the largest degradation on compositional tasks (3.1\% on OneIG-Bench, 2.5\% on CompBench), as off-topic edits undermine maintaining compositional constraints. Removing the minimal visual changes filter most significantly hurts multi-turn editing (1.16 points on ImgEdit), demonstrating that learning meaningful incremental progress is critical for sustained interactions. For visual reasoning (MIRA), removing the quality regression filter has the largest impact (1.5\%), as learning from degraded trajectories impairs converging toward correct answers. Different quality dimensions matter for different capabilities; effective test-time scaling requires curating data along multiple axes.

\begin{table}[t]
\centering
\small
 \resizebox{0.9\linewidth}{!}{%
\begin{tabular}{lcccc}
\toprule
\textbf{Data Configuration} & \textbf{OneIG} & \textbf{CompBench} & \textbf{ImgEdit} & \textbf{MIRA} \\
& \textbf{Align (\%)} & \textbf{Overall (\%)} & \textbf{Score} & \textbf{Acc (\%)} \\
\midrule
Full curated dataset & 84.3 & 98.8 & 4.26 & 11.5 \\
w/o Quality regression filter & 82.5 (-1.8) & 97.5 (-1.3) & 3.30 (-0.96) & 10.0 (-1.5) \\
w/o Relevance filtering & 81.2 (-3.1) & 96.3 (-2.5) & 3.60 (-0.66) & 10.3 (-1.2) \\
w/o Min. visual changes filter& 83.5 (-0.8) & 98.0 (-0.8) & 3.10 (-1.16) & 10.9 (-0.6) \\
\bottomrule
\end{tabular}%
}
\caption{\textbf{Data quality ablation.} Impact of removing individual curation filters (Sec.~\ref{sec:data_collection}) from the training data pipeline.}
\label{tab:data_quality}
\end{table}

%% file: tables/mira.tex
\begin{table}[t]
\centering
\small
\setlength{\tabcolsep}{3pt}
\begin{tabular}{l|cccc|c}
\toprule
\textbf{Method} & \textbf{EG}$\uparrow$ & \textbf{PBR}$\uparrow$ & \textbf{ASLP}$\uparrow$ & \textbf{CT}$\uparrow$ & \textbf{Overall}$\uparrow$ \\
\midrule
GPT-5 \citep{openai2025gpt5} & \textbf{14.5} & \textbf{29.9} & 10.8 & \textbf{17.9} & \textbf{16.5} \\
Qwen2.5-VL (72B) \citep{Qwen2.5} & \textbf{14.5} & 21.7 & \textbf{11.1} & 8.6 & 13.1 \\
Bagel \citep{deng2025bagel} & 9.7 & 7.9 & 3.5 & 12.3 & 7.5 \\
Bagel+CoT \citep{deng2025bagel} & 8.0 & 9.5 & 4.8 & 14.2 & 9.2 \\
\midrule
\methodname & 12.5 & 11.2 & 6.1 & 16.3 & 11.5 \\
\bottomrule
\end{tabular}%
\caption{\textbf{Multimodal reasoning, MIRA}, with direct input. We report results across four reasoning categories: EG (Geometry), PBR (Physics), ASLP (Puzzles), and CT (Causal), along with the overall average score.}
\label{tab:mira_comparison}
\label{tab:mira_results}
\end{table}

%% file: sec/6_conclusion.tex
\section{Conclusion}
\label{sec:conclusion}

We have presented a unified approach to multimodal chain-of-thought test-time scaling that extends inference-time compute from text-only reasoning to models handling both visual understanding and generation, establishing a paradigm that benefits both generation and comprehension across modalities. Our key contributions—an agentic framework that induces cognitive behaviors (verification, subgoal decomposition, content memory), budget forcing for beyond-training generalization, and evidence that sequential reasoning outperforms parallel sampling—yield substantial gains across compositional generation, multi-turn editing, and visual reasoning, demonstrating that iterative refinement through explicit reasoning unlocks significant performance improvements on complex multimodal tasks.

%% file: sec/7_supp.tex
\begin{figure*}[!t]
\centering
\includegraphics[width=\linewidth]{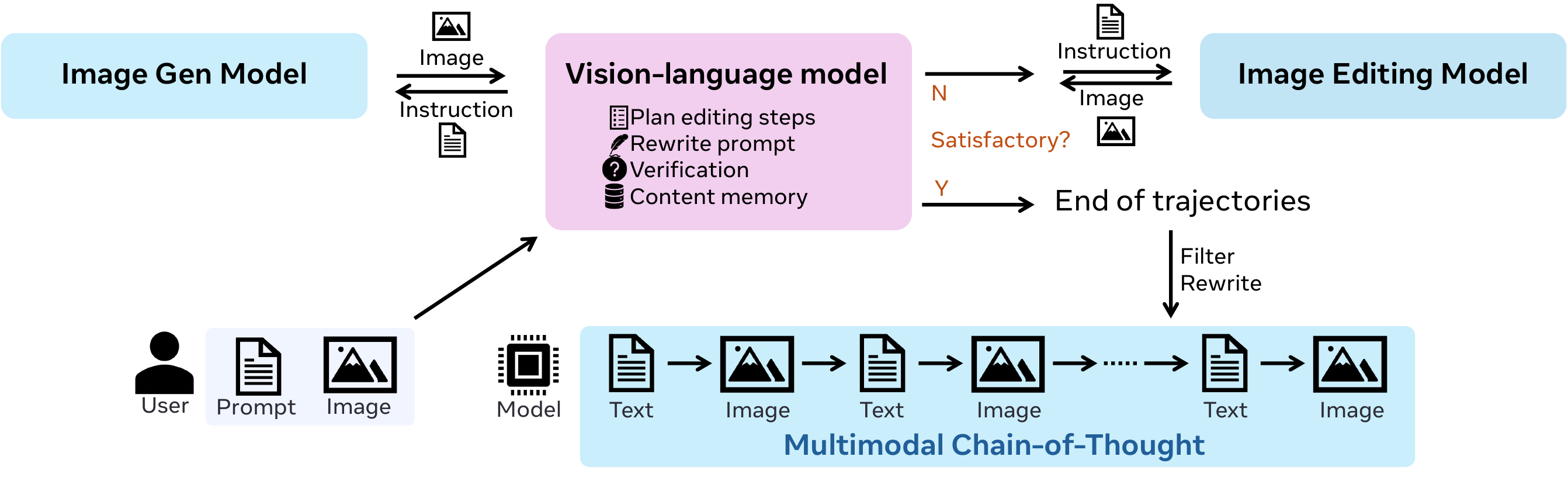}
\caption{\textbf{Data synthesis pipeline architecture.} Three model roles coordinate via information flows: Image Gen Model produces initial images, Vision-language model verifies image and performs planning/prompt rewriting with content memory, Image Editing Model applies refinements. Trajectories loop until satisfied, producing interleaved text-image chain-of-thought data.}
\label{fig:pipeline_overview}
\end{figure*}

\section{Data Synthesis Pipeline}
\label{supp:data}

We provide implementation details for our automated data collection pipeline that generates 12K multimodal chain-of-thought training trajectories.

\subsection{Pipeline Architecture}

The agentic framework coordinates three model roles in an iterative loop (Fig.~\ref{fig:pipeline_overview}): \textbf{Image Gen Model} produces initial images from user prompts, \textbf{Vision-language Model} evaluates satisfaction and performs verification with content memory and subgoal decomposition, and \textbf{Image Editing Model} applies refinements based on VLM planning. This loop continues until the VLM determines the image satisfies all requirements, producing interleaved text-image chain-of-thought trajectories that are filtered for quality.

\subsection{Model Components}

\noindent \textbf{Prompt generation:} Llama-4-Scout-17B-16E generates 20K diverse prompts covering compositional attributes, spatial relations, and multi-object generation tasks based on T2I-CoReBench.

\noindent \textbf{Image generation:} Flux Pro produces initial images from prompts. For complex prompts, Qwen3-VL decomposes prompts into subgoals and executes the first step in initial generation.

\noindent \textbf{Verification:} Qwen3-VL evaluates whether images satisfy prompts. If not, it generates explicit chain-of-thought reasoning, identifying deficiencies, planning improvements, and specifying editing instructions.

\noindent \textbf{Editing:} Flux Kontext or Qwen-Image-Edit applies editing instructions based on VLM planning.

\subsection{Example Trajectory}

\begin{figure*}

\centering
\includegraphics[width=\linewidth]{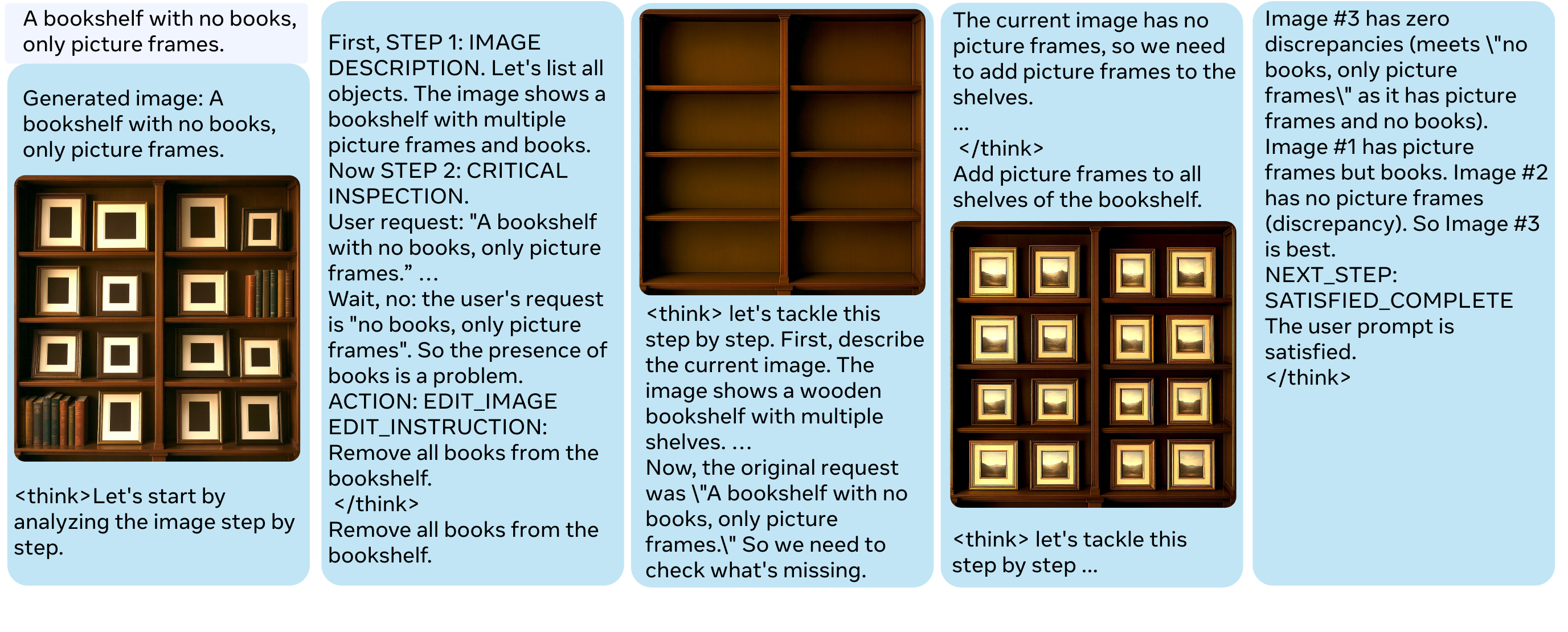}
\caption{\textbf{Detailed chain-of-thought trajectory demonstrating cognitive behaviors.} This bookshelf generation example shows the model's explicit reasoning through \texttt{<think>} blocks across three refinement rounds. \textbf{Verification:} the model identifies that books are present when the prompt specifies ``no books, only picture frames.'' \textbf{Subgoal decomposition:} the model breaks the correction into sequential steps—first removing books, then adding picture frames. \textbf{Content memory:} the model explicitly references and compares Images \#1, \#2, and \#3 to track cumulative progress. The reasoning demonstrates how chain-of-thought enables iterative self-correction through explicit evaluation and planning.}
\label{fig:verification}
\end{figure*}

Fig.~\ref{fig:verification} shows a concrete bookshelf generation trajectory demonstrating the three cognitive behaviors induced by our framework. The VLM performs \textbf{verification} by identifying that books are present when the prompt specifies ``no books, only picture frames.'' It exhibits \textbf{subgoal decomposition} by breaking the correction into sequential steps—first removing books, then adding frames. Finally, it demonstrates \textbf{content memory} by explicitly referencing and comparing Images \#1, \#2, and \#3 to track cumulative progress across refinement rounds.

\subsection{Training Data Statistics}

After quality filtering, we obtain 12K trajectories with the following characteristics: training trajectories average \textbf{3.6 refinement rounds} (range 1-8 rounds). Training on this data requires 700 H100 GPU hours.

\subsection{VLM Prompt Design}

The vision-language model uses a structured prompt template (Table~\ref{tab:prompt_verification}) to induce cognitive behaviors during data synthesis. The prompt guides the VLM through three steps: (1) detailed image description with explicit object counts and spatial relationships, (2) comparison analysis against user requirements with reflection on previous attempts, and (3) decision making between editing, backtracking, or completion. This structured reasoning naturally produces trajectories exhibiting verification, subgoal decomposition, and content memory.

\begin{table}[!ht]\centering
\begin{minipage}{0.7\textwidth}\vspace{0mm}
    \centering
    \begin{tcolorbox}
        \centering
        \begin{tabular}{p{0.65\textwidth}}
        \hspace{1mm}
        \begin{minipage}{0.63\textwidth}
        \small
        \texttt{You are an intelligent and honest image evaluation agent.}\\
        \\
        \texttt{ORIGINAL USER REQUEST: \{user\_prompt\}}\\
        \texttt{[Previous images information with satisfied/TODO features]}\\
        \\
        \texttt{STEP 1 - IMAGE DESCRIPTION:}\\
        \texttt{First, describe what you see in the current image in detail:}\\
        \texttt{- List ALL objects present with exact counts}\\
        \texttt{- Describe their positions and spatial relationships}\\
        \texttt{- Note colors, materials, lighting, and style}\\
        \texttt{- Describe the overall scene composition}\\
        \\
        \texttt{STEP 2 - COMPARISON ANALYSIS:}\\
        \texttt{Compare your image description with the user request:}\\
        \texttt{1. COUNT all objects explicitly}\\
        \texttt{2. CHECK spatial relationships}\\
        \texttt{3. VERIFY colors, materials, and other specific details}\\
        \texttt{4. IDENTIFY correct objects to retain and wrong objects to remove}\\
        \texttt{5. REFLECT on previous attempts - making progress or stuck?}\\
        \texttt{6. If instruction failed multiple times, try simpler language}\\
        \\
        \texttt{STEP 3 - DECISION:}\\
        \texttt{Choose ONE action:}\\
        \\
        \texttt{ACTION: EDIT\_IMAGE}\\
        \texttt{EDIT\_INSTRUCTION: [5-18 word instruction, focus on ONE change]}\\
        \texttt{SATISFIED: [features matching request with counts]}\\
        \texttt{TODO: [features still needed with counts]}\\
        \\
        \texttt{ACTION: BACKTRACK\_TO\_IMAGE}\\
        \texttt{BACKTRACK\_TO: [image number, e.g., "Image \#2"]}\\
        \\
        \texttt{ACTION: SATISFIED\_COMPLETE}\\
        \texttt{SATISFIED: [all requirements met with verification]}
        \end{minipage}
        \end{tabular}
    \end{tcolorbox}
    \vspace{-2mm}
    \caption{\textbf{VLM verification and planning prompt.} Structured template guiding the vision-language model through image description, comparison analysis, and action decision. This three-step reasoning naturally induces verification, subgoal decomposition, and content memory behaviors during trajectory generation.}
    \label{tab:prompt_verification}
    \end{minipage}
    \vspace{-2mm}
\end{table}

\section{Additional Qualitative Results}
\label{supp:qualitative}

\begin{figure*}[!t]
\centering
\includegraphics[width=\linewidth]{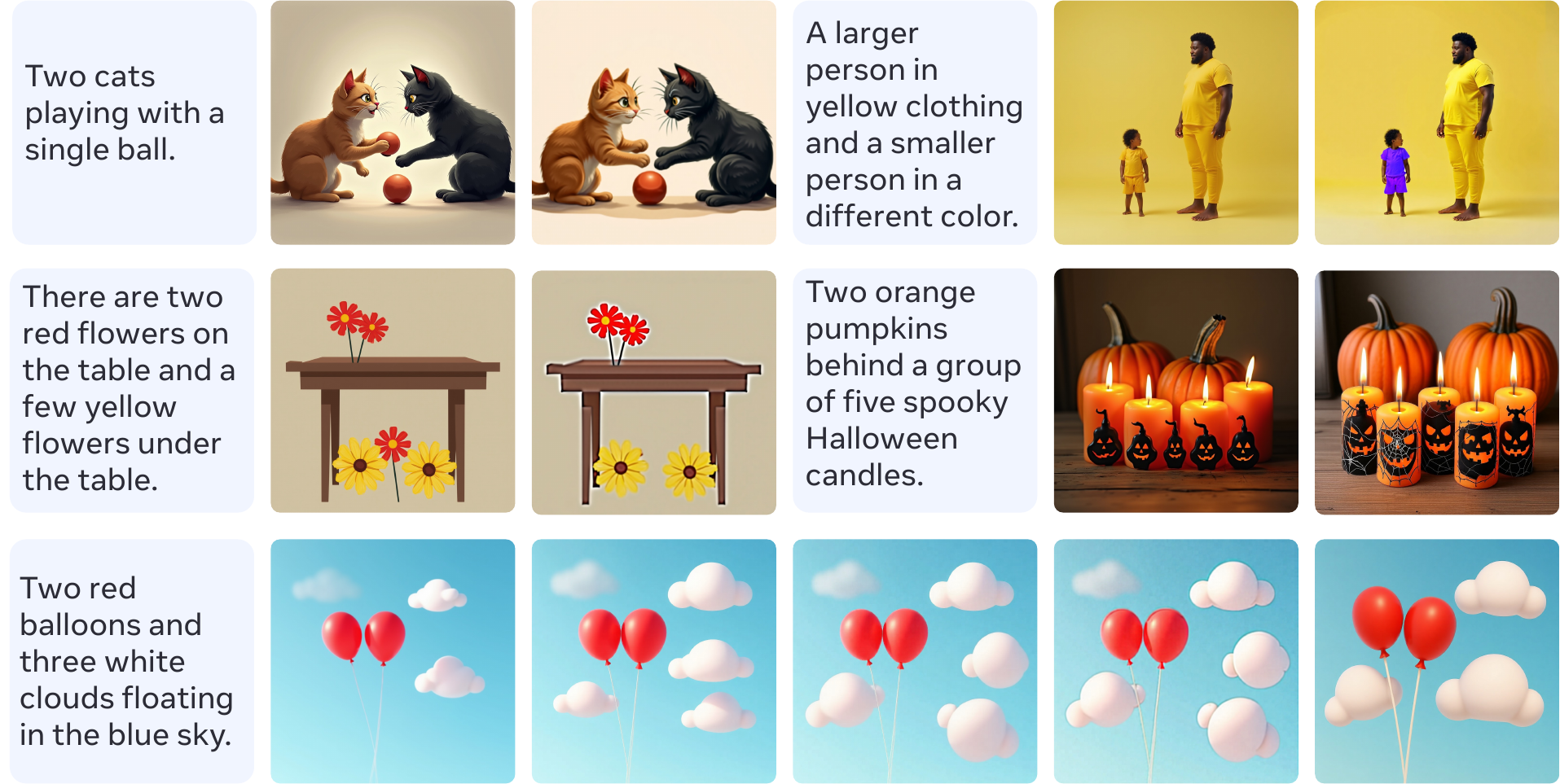}
\caption{\textbf{Qualitative examples of chain-of-thought test-time scaling.} Representative trajectories showing progressive refinement across different tasks and computational budgets. Examples demonstrate how explicit chain-of-thought reasoning enables the model to iteratively improve compositional generation.}
\label{fig:visual_supp}
\end{figure*}
The VLMs produce interleaved text and image tokens with explicit thinking tokens. The reflection step generates detailed reasoning about \emph{why} outputs fall short and \emph{how} to improve them, rather than simply issuing new instructions.

Additional qualitative examples are shown in Fig.~\ref{fig:visual_supp}, covering representative trajectories across different task types and computational budgets.

\section{Generalization Preservation}
\label{supp:forgetting}

Fine-tuning on 12K reasoning-heavy trajectories does not cause catastrophic forgetting of the base Bagel model's general capabilities. We compare Bagel before and after fine-tuning on our data (without test-time scaling at inference): the fine-tuned model achieves 0.783 alignment on OneIG-Bench (vs.\ 0.764 for vanilla Bagel) and 2.26 on ImgEdit (vs.\ 1.31), indicating that the chain-of-thought training data improves rather than degrades the base model's instruction-following capabilities even without multi-round inference.

\section{Scaling Beyond $C{=}10$}
\label{supp:scaling_limit}

We cap evaluation at $C{=}10$ due to GPU memory constraints. We observe that image quality collapses when editing rounds produce minimal visual changes (LPIPS $< 0.03$ between consecutive images), as accumulated autoregressive noise degrades fidelity. Such non-improving editing steps are infrequent, so scaling remains effective up to $C{=}10$. Beyond this point, we expect TTS performance to saturate or degrade once quality collapse dominates. The exact inflection point depends on the base generation and editing model capabilities. Potential mitigations include: (a) perceptual thresholding to skip rounds with minimal changes, (b) ``reset'' rounds that regenerate from scratch using accumulated reasoning, and (c) adaptive noise scheduling to counteract quality degradation.

\section{Failure Analysis}
\label{supp:failure}
\subsection{Failure Cases}
\label{subsec:failure_cases}

\begin{figure*}[!t]
\centering
\includegraphics[width=\linewidth]{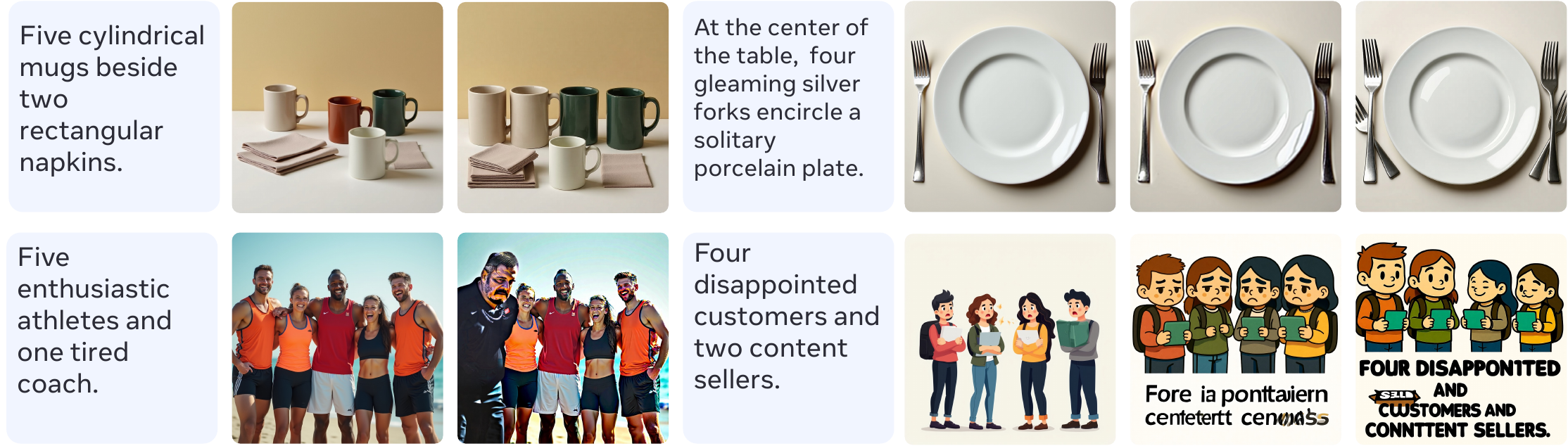}
\caption{\textbf{Representative failure modes.} \textbf{Example 1:} Compositional constraints with precise object counts and spatial arrangements (napkin count). \textbf{Example 2:} Complex spatial relationships requiring specific geometric configurations (forks encircling plate). \textbf{Examples 3,4:} Layout change from the intermediate images (people count).}
\label{fig:failure}
\end{figure*}

Despite strong performance, our approach exhibits limitations in specific scenarios. First, tasks requiring precise physical reasoning or fine-grained spatial relationships occasionally fail, as iterative refinement may struggle to correct fundamental physics violations or attribute binding errors inherited from the base generation/editing models (e.g., incorrect leash-dog assignment or wrong helmet placement and sizes in Fig.~\ref{fig:comparison}). Second, we observe occasional degradation loops where reflection incorrectly identifies non-existent issues, leading to unnecessary edits that harm quality rather than improve it—a verification hallucination bottleneck particularly evident when the VLM's verification capabilities are insufficient to accurately assess subtle visual attributes. Third, extremely complex compositional prompts with many interacting constraints can lead to subgoal conflicts during decomposition, where satisfying one constraint inadvertently violates another. Finally, test-time scaling cannot overcome fundamental capability gaps in the base model; if the underlying diffusion or VLM components lack certain semantic understanding, additional inference compute provides diminishing returns. Representative examples are shown in Fig.~\ref{fig:failure}. These failure modes suggest directions for future work, including more robust verification mechanisms, physics-aware refinement strategies, and constraint satisfaction planning. We further discuss scaling limits beyond $C{=}10$ in Sec.~\ref{supp:scaling_limit}.

\noindent \textbf{Limitations.} While our approach demonstrates strong results, test-time scaling inherently requires additional computational resources at inference. Future work should explore more efficient reflection mechanisms and adaptive budget allocation strategies that minimize computational overhead while preserving quality gains.

\noindent \textbf{Future directions.} Promising directions include extending our approach to additional modalities (audio, video), augmenting reflection with explicit physical reasoning to enforce implicit constraints (e.g., object sizing, perspective, occlusion), investigating reinforcement learning from human feedback to further improve reflection quality, and exploring how test-time scaling interacts with other inference-time techniques such as self-consistency and verifier-guided generation.